\documentclass{article}
\usepackage{imav}
\usepackage{tikz,pgfplots}
\usepackage{aircraftshapes}
\usepackage{times}
\usepackage{graphicx}
\usepackage{amsmath} 
\usepackage{amssymb}  
\usepackage{comment}
\usepackage[]{algorithm2e}
\usepackage{caption}
\usepackage{subcaption}

\usepackage[square,numbers]{natbib}
\usetikzlibrary{calc, arrows.meta}

\newcommand{\dfb}{\stackrel{\Delta}{=}}
\newcommand{\R}{\ensuremath{\mathbb R}}

\title{Distributed circular formation flight of fixed-wing aircraft with Paparazzi autopilot}
\author{Hector Garcia de Marina\thanks{hector.garcia-de-marina@enac.fr} \quad Gautier Hattenberger\thanks{gautier.hattenberger@enac.fr} \\ ENAC, University of Toulouse, F-31055, France}

\begin{document}

\maketitle
\thispagestyle{empty} 

\begin{abstract}
In this paper we introduce the usage of guidance vector fields for the coordination and formation flight of fixed-wing aircraft. In particular, we describe in detail the technological implementation of the formation flight control for a fully distributed execution of the algorithm by employing the open-source project Paparazzi. In this context, distributed means that each aircraft executes the algorithm on board, each aircraft only needs information about its neighbors, and the implementation is straightforwardly scalable to an arbitrary number of vehicles, i.e., the needed resources such as memory or computational power not necessarily scale with the number of total aircraft. The coordination is based on commanding the aircraft to track circumferences with different radii but sharing the same center. Consequently, the vehicles will travel different distances but with the same speeds in order to control their relative angles in the circumference, i.e., their orbital velocities. We show the effectiveness of the proposed design with actual formation flights during the drone parade in IMAV2017.
\end{abstract}

\section{Introduction}
\label{section:intro}
Unmanned Aerial Vehicles (UAVs) have emerged as powerful platforms for, among others, atmospheric research, intensive agriculture and surveillance. A key aspect of the usage of these vehicles is to make affordable and accessible certain tasks within these fields. In particular, the technologies around these aerial vehicles, such as batteries, propulsion systems, construction materials and their associated maintenance, have developed an interesting performance for the actual cost. This fact makes interesting the usage of several of these vehicles in a cooperative way in order to enhance tasks that were previously performed by a single aircraft. For example, we can perform missions more efficiently by splitting the payload of one aircraft, such as antennas \cite{jin2010gnss} or different kind of surveillance cameras \cite{maza2011experimental} to two or more aircraft. Furthermore, we can also employ the coordination of several aircraft for the estimation of the wind without employing expensive sensors \cite{mayer2012no}. Another interesting and different usage of fleets of UAVs are related with performance shows such as the suggested outdoor parade contest of IMAV2017\footnote{http://www.imavs.org/2017}.

There exist a large number of algorithms in literature related with the formation control of vehicles \cite{oh2015survey}. However, most of them consider the dynamics of the vehicle as kinematic points, which is a very restrictive requirement for fixed-wing aircraft that can be seen as an unicycle. Formation control algorithms dealing with this kind of dynamics can still be found in the literature \cite{marshall2006pursuit,sepulchre2008stabilization,uni2017}. However, most of them do not consider desirable requirements such as vehicles with constant speed, e.g., with motors operating constantly at their nominal conditions, or it is not clear how to combine such algorithms with trajectory tracking, which is quite important in order to guarantee the confinement of the UAVs in their allowed airspace.

This paper aims to show how the algorithm proposed in \cite{iros2017} has been implemented in a fully distributed way by employing the open-source project Paparazzi. The algorithm proposes a technique based on guidance vector fields for trajectory tracking \cite{YuriCS} in order to achieve the coordination of multiple fixed-wing aircraft flying at constant (and equal) speeds. In particular, we focus on circular trajectories but it is also applicable to any other closed trajectories. The main idea relies on controlling the traveled distance by the aircraft by commanding them to track a smaller or bigger radius with respect to the desired circle. Consequently, the vehicles will travel different distances but with the same speeds in order to control their relative angles in the circumference, i.e., the algorithm controls their orbital velocities. The aircraft are not always placed on the commanded circles but far away from them, specially in the beginning of the algorithm where the initial positions of the vehicles could be arbitrary. Nevertheless, the proposed strategy is feasible because of the exponential convergence property to the desired trajectory of the guidance algorithm \cite{YuriCS}. Therefore if the convergence of the formation flight algorithm is slow enough, then the whole cascaded system (trajectory tracking and formation control) will converge to the desired coordinated formation flight.

We organize this paper as follows. The Section \ref{sec: gvf} reviews the concept of vector field for tracking smooth trajectories. Then, we explain in Section \ref{sec: fc} how to manipulate such vector fields in order to achieve the desired circular formation flight. We continue in Section \ref{sec: imp} showing how the algorithm is fully distributed and executed in the open-source Paparazzi autopilot \cite{hattenberger2014using}. In particular, in Section \ref{sec: ff} we show as an example a simulation of the formation flight parade that will be performed during IMAV 2017. We end the paper with some conclusions in Section \ref{sec: con}.

\section{Guidance vector field for tracking smooth trajectories}
\label{sec: gvf}
\subsection{Fixed-wing aircraft's model}
Consider for the \emph{unit speed} fixed-wing aircraft the following nonholonomic model in 2D
\begin{equation}
\begin{cases}
	\dot p &= m(\psi) \\
	\dot\psi &= u_{\psi},
\end{cases}
	\label{eq: pdyn}
\end{equation}
where $p = \begin{bmatrix}p_x & p_y\end{bmatrix}^T$ is the Cartesian position of the vehicle, $m = \begin{bmatrix}\cos(\psi) & \sin(\psi)\end{bmatrix}^T$ with $\psi\in(-\pi, \pi]$ being the attitude \emph{yaw} angle\footnote{For our setup, the yaw angle and heading angle can be considered equal due to the absence of wind.} and $u_{\psi}$ is the control action that will make the aircraft to turn. If we consider that the altitude of the vehicle is kept constant and its pitch angle is close to zero, then the control action $u_{\psi}$ corresponds to the following bank angle $\phi$ in order to have a coordinated turn 
\begin{equation}
	\phi = \operatorname{arctan}\frac{u_{\psi}}{g},
	\label{eq: uphi}
\end{equation}
where $g$ is the gravity acceleration.

\subsection{Trajectory tracking}
We have chosen the algorithm proposed in \cite{YuriCS} for the task of tracking circular trajectories since it has been successfully validated in real flights \cite{de2016guidance}. One interesting property of the chosen algorithm is that the local exponential converge to the desired path is guaranteed. This property help us to support the convergence of the \emph{high level} formation control algorithm under the argument of \emph{slow-fast} dynamical systems \cite{iros2017}.

\begin{figure}
	\begin{tikzpicture}
	\draw[fill=gray,cm={cos(-55),-sin(-55),sin(-55),cos(-55),(1.15,0)}](0,1.3)--(0,0.7)--(1,1)--(0,1.3);
	\draw[dashed] (3,-4) ++(75:5.4) arc (75:130:5.4) node[at start, xshift=35]{$\varphi(p^*) = e > 0$};
		\draw[dashed] (3,-4) ++(75:2.3) arc (75:130:2.3) node[at start, xshift=25]{$\varphi(p) < 0$};
		\draw (3,-4) ++(75:3.3) arc (75:130:3.3) node[at start, xshift=35]{$\mathcal{P}:=\varphi(p) = 0$};;
	\draw[-{Latex[length=8, width=8]}] (2.8, -4) -- (3.8, -4);
		\draw[-{Latex[length=8, width=8]}] (2.8, -4) node[left]{HOME} -- (2.8, -3);
	\draw[-{Latex[length=8, width=8]}] (2.8, -4) -- (0.5,0.8) node[pos=0.3, above, xshift=5]{$p^*$};
	\draw[-{Latex[length=8, width=8]}] (0.5,0.8) -- (1.75, 1.3) node[pos=1, below,yshift=-3]{$\tau$};
	\draw[-{Latex[length=8, width=8]}] (0.5,0.8) -- (0, 2.05) node[pos=1, above]{$n$};
	\draw[-{Latex[length=8, width=8]}] (0.5,0.8) -- (1.3, -0.85) node[pos=0.8, left]{$-k_een$};
		\draw[-{Latex[length=8, width=8]}, color=red] (0.5,0.8) -- (3.05-1.5, 1.3-1.1) node[pos=0.7, above]{$\hat{\dot p}_d$};
	\draw[-{Latex[length=8, width=8]}] (0.5,0.8) -- (1.33,2) node[pos=1.15]{$\dot p$};
	\draw[dashed] (1, 1.5) -- (1.5, 1.5);
	\draw[dashed] (1.3, -0.85) -- (3.05-0.5, 1.3-0.85-0.8);
	\draw[dashed] (1.75, 1.3) -- (3.05-0.5, 1.3-0.85-0.8);
	\draw (1.45,1.5) arc (0:30:1) node[pos=-0, xshift=5, right, above]{$\psi$};
\end{tikzpicture}
	\caption{The direction to be followed by the UAV at the point $p^*$ with respect to \emph{HOME} for converging to $\mathcal{P}$ is given by $\hat{\dot p}_d$. The tangent and normal vectors $\tau$ and $n$ are calculated from $\nabla\varphi(p^*)$. The error \emph{distance} $e$ is calculated as $\varphi(p^*)$.}
	\label{fig: ilus}
\end{figure}
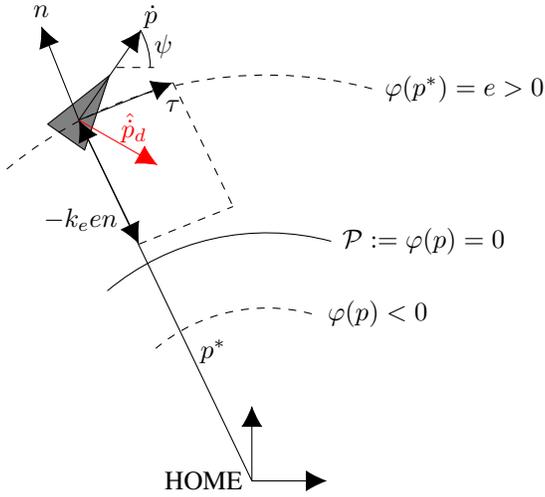

\begin{figure}[h]
\centering
	\includegraphics[width=1.0\columnwidth]{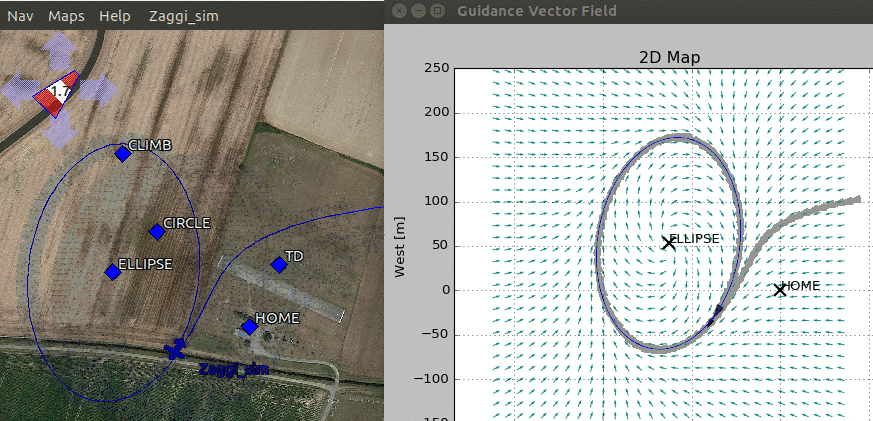}
	\caption{Example of the directions $\hat{\dot p}_d$ given by the vector field (\ref{eq: gvf}) for tracking an ellipsoidal track. Screenshot taken from the real time \emph{gvf app} available in Paparazzi.}
	\label{fig: sin}
\end{figure}

Consider the circular path $\mathcal{P}$ described by
\begin{equation}
	\mathcal{P}:= \{p \, : \varphi(p) := p_x^2 + p_y^2 - r^2 = 0\}.
	\label{eq: P}
\end{equation}
Clearly the function $\varphi : \mathbb{R}^2 \to \mathbb{R}$ belongs to the $C^2$ space and it is \emph{regular} everywhere excepting at the origin, i.e.,
\begin{equation}
	\nabla\varphi(p) = \begin{bmatrix}2p_x \\ 2p_y \end{bmatrix} \neq 0 \iff p\in \mathbb{R}^2 \setminus 0.
	\label{eq: reg}
\end{equation}
The trajectory tracking algorithm employs the level sets $e(p)\dfb\varphi(p)$ for the notion of \emph{error distance} between the aircraft and $\mathcal{P}$. In particular, for circular trajectories a positive level set corresponds to an \emph{expanded version} of the desired circle $\mathcal{P}$, while a negative level set corresponds to a \emph{contracted version}. Note that the domain of the error distance is $e\in[-r^2, \infty)$. We define by $n(p) := \nabla \varphi(p)$ the normal vector to the curve corresponding to the level set $\varphi(p)$ and the tangent vector $\tau$ at the same point $p$ is given by the rotation
\begin{equation}
	\tau(p) = En(p) = \begin{bmatrix}2p_y \\ -2p_x\end{bmatrix}, \quad E=\begin{bmatrix}0 & 1 \\ -1 & 0\end{bmatrix}. \nonumber
\end{equation}
Note that $E$ will determine in which direction $\mathcal{P}$ will be tracked, which is done by following the direction at each point $p$ given by of the vector field
\begin{equation}
	\dot p_d(p) \dfb \tau(p) - k_e e(p)n(p) = 2\begin{bmatrix}p_y - k_eep_x \\ p_x -k_eep_y \end{bmatrix},
	\label{eq: gvf}
\end{equation}
where $k_e\in\R^+$ is a gain that defines how \emph{aggressive} is the vector field. An example of the construction of the guidance vector field (\ref{eq: gvf}) is shown in Figure \ref{fig: ilus}, and its visualization in Paparazzi for tracking an ellipsoidal trajectory is shown in Figure \ref{fig: sin}.

Let us define $\hat x$ as the unit vector constructed from the nonzero vector $x$. The vector field (\ref{eq: gvf}) is successfully tracked if we apply the following control action to (\ref{eq: uphi}) \cite{YuriCS,de2016guidance} 
\begin{align}
	u_\psi &= -\left(E\hat{\dot p}_d\hat{\dot p}_d^TE\left((E-k_ee)H(\varphi)\dot p - k_en^T\dot pn\right)\right)^TE\frac{\dot p_d}{||\dot p_d||^2} \nonumber \\
	&+ k_d\hat{\dot p}^TE\hat{\dot p}_d,
	\label{eq: ui}
\end{align}
	where $H(\cdot)$ is the Hessian operator, i.e., from (\ref{eq: reg}) we have that $H = \begin{bmatrix}2 & 0 \\ 0 & 2\end{bmatrix}$ and $k_d\in\R^+$ is a positive gain that determines how fast the vehicle converges to the guidance vector field. We can clearly identify two terms in the addition in (\ref{eq: ui}). The first term is a feedforward component and makes the aircraft to stay on the guidance vector field (\ref{eq: gvf}) while the second term makes the vehicle to converge to the guidance vector field in case that the vehicle is not aligned with it.

\section{Guidance vector field as a coordinating tool}
\label{sec: fc}
Consider a team of $n$ aircraft traveling all of them at the same speed. The main objective of this paper is to show how to make them rendezvous without actuating on their traveling velocities. The rendezvous will happen at the same time that the team is traveling over a desired circular trajectory $\mathcal{P}$. We will see that this is possible to achieve by controlling the traveled distance of the aircraft around $\mathcal{P}$.

\subsection{Circular trajectory}
We summarize in this subsection the formation control algorithm presented in \cite{iros2017}. Consider that the center of $\mathcal{P}$ is at the origin as in (\ref{eq: P}). Let us define the phase of the aircraft as $\theta = \operatorname{atan2}(p_y,p_x)$ where
\begin{equation}
\operatorname{atan2}(y,x) =
\begin{cases}
\arctan(\frac y x) &\text{if } x > 0, \\
\arctan(\frac y x) + \pi &\text{if } x < 0 \text{ and } y \ge 0, \\
\arctan(\frac y x) - \pi &\text{if } x < 0 \text{ and } y < 0, \\
+\frac{\pi}{2} &\text{if } x = 0 \text{ and } y > 0, \\
-\frac{\pi}{2} &\text{if } x = 0 \text{ and } y < 0, \\
\text{undefined} &\text{if } x = 0 \text{ and } y = 0.
\end{cases}
\end{equation}

\begin{figure}
	\centering
	\begin{tikzpicture}
	\draw circle (3);
	\draw [dashed] circle (3.4);
	\draw [dashed] circle (2.6);
	\node [aircraft top,fill=black,minimum width=0.5cm,rotate=0] at (0,3.4) {};
	\node [aircraft top,fill=black,minimum width=0.5cm,rotate=90, color=red] at (-2.6,0) {};
	\draw (0,0) -- (-2,0);
	\draw (0,0) -- (0,2);
		\draw (-1,0) arc (180:90:1) node[pos=0.3, above, xshift=-5] {$z_{1\theta}$};
		\draw [->, thick] (0,0) -- (2.6,0) node[pos=0.5, above] {$r-k_\theta z_{1\theta}$};
		\draw [->, thick] (0,0) -- (0, -3.4)  node[pos=0.5, xshift=25] {$r+k_\theta z_{1\theta}$};
	\end{tikzpicture}
	\caption{Red and black airplane want to fly over $\mathcal{P} := x^2 + y^2 - r^2 =0$ (solid circle) with the same phase $\theta_1 = \theta_2$, i.e., $z_{1\theta} := \theta_1 - \theta_2 = 0$. Red airplane travels less distance if it tracks a smaller radius (negative level set with respect to $\mathcal{P}$), therefore its angular velocity with respect to the center of $\mathcal{P}$ is bigger than the angular velocity of black airplane. Once $z_{1\theta} = 0$, both aircraft are tracking $\mathcal{P}$. Setting the positive gain $k_\theta$ will tune the convergence time for the rendezvous.}
	\label{fig: syncc}
\end{figure}
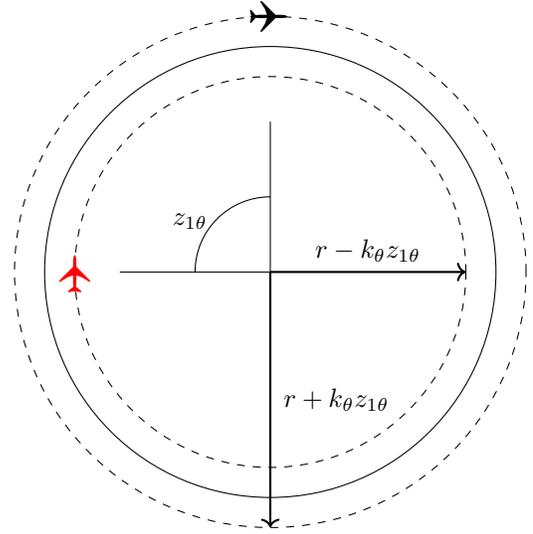

In a circular trajectory we are interested in controlling the different inter-vehicle phases, e.g., between aircraft $1$ and $2$ we could define $z_1 := \theta_1 - \theta_2$. For a general case, the relationships between neighbors are described by an undirected graph $\mathbb{G} = (\mathcal{V}, \mathcal{E})$ with the vertex set $\mathcal{V} = \{1, \dots, n\}$ and the ordered edge set $\mathcal{E}\subseteq\mathcal{V}\times\mathcal{V}$. The set $\mathcal{N}_i$ of the neighbors of agent $i$ is defined by $\mathcal{N}_i\dfb\{j\in\mathcal{V}:(i,j)\in\mathcal{E}\}$. We define the elements of the incidence matrix $B\in\R^{|\mathcal{V}|\times|\mathcal{E}|}$ for $\mathbb{G}$ by
\begin{equation}
	b_{ik} \dfb \begin{cases}+1 \quad \text{if} \quad i = {\mathcal{E}_k^{\text{tail}}} \\
		-1 \quad \text{if} \quad i = {\mathcal{E}_k^{\text{head}}} \\
		0 \quad \text{otherwise}
	\end{cases}.
	\label{eq: B}
\end{equation}
Note that the $i$'th row of $B$ denotes for the $i$'th vehicle and the $k$'th column denotes for the link $\mathcal{E}_k$. Define $\Theta\in\mathbb{R}^{|\mathcal{E}|}$ as the stacked vector of aircraft phases, then one can calculate
\begin{equation}
	z = B^T\Theta,
\end{equation}
as the stacked vector of all the available inter-vehicle phases $z_k(t)$ to be controlled. In particular, for rendezvous, we are interested in driving the signal $z(t)$ to zero. Depending on which level set of $\mathcal{P}$ an aircraft is following, it will travel with a different angular velocity $\dot\theta_i$ with respect to the center of $\mathcal{P}$, allowing us to control the evolution of the different $z_k$'s. In order to track different level sets of $\mathcal{P}$ we introduce a control signal $^iu_r$ to (\ref{eq: P}) for the $i$'th aircraft as
\begin{equation}
	\varphi(p) := p_x^2 + p_y^2 - (r+\,{^iu_r})^2 = 0,
	\label{eq: radiuseq}
\end{equation}
and we apply a consensus algorithm
\begin{equation}
	^iu_r = k_r B_iz,
	\label{eq: ur}
\end{equation}
where $B_i$ stands for the $i$'th row of the incidence matrix $B$ and $k_r\in\R^+$ is a positive gain. The results in \cite{iros2017} guarantees the exponential stability of the equilibrium at $\Theta = 0$ if $\mathcal{G}$ does not contain any cycles, e.g., a spanning tree topology. An example for two aircraft can be found in Figure \ref{fig: syncc}.

\section{Implementation in Paparazzi}
\label{sec: imp}
We discuss in this section the implementation of the formation flight algorithm in the open-source project Paparazzi. We start by rewriting the control action (\ref{eq: ur}) as follows
\begin{equation}
	^iu_r = k_r \sum_{j\in\mathcal{N}_i} (\theta_i - \theta_j),
	\label{eq: urpapa}
\end{equation}
therefore the $i$'th aircraft only needs to collect the phases $\theta_j$ from its neighbors. We do this process of collection by inter-vehicle communication, i.e., each aircraft calculates its phase $\theta_i$ and transmits it to its neighbors. Note that since we are controlling inter-vehicle phases, the aircraft can control their relative phases while following circumferences with different centers for $\mathcal{P}$, although obviously the rendezvous will not take place in such a situation.

We write the software responsible of running the algorithm in Paparazzi as a module\footnote{https://wiki.paparazziuav.org/wiki/Modules}. A module allows one to add new code in a flexible way with initialization, periodic and event functions without modifying the main autopilot loop. Each aircraft will run exactly the some code without any particular modification. This will allow the scaling up of the number of aircraft without any associated penalty. In order to run the algorithm each aircraft is required in their flight plan to be tracking a circle with the guidance vector field algorithm\footnote{https://wiki.paparazziuav.org/wiki/Module/guidance\_vector\_field}.

The module employs the new functionalities of Paparazzi Link v2.0\footnote{Paparazzi Link is a messages toolkit (message definition, code generators, libraries) to be used with Paparazzi and compatible systems}. This communication layer allows the inter-vehicle communication, i.e., air to air, without intervention of the Ground Control Station. Each aircraft in Paparazzi has an unique identification ID (uint8) that will be employed for each aircraft $i$ in (\ref{eq: urpapa}). The formation flight module runs two independent processes. The first process keeps updated a table once a message from a neighbor is received. This table shown in Table \ref{table: t} has information about the IDs of the neighbors, their latest received $\theta_j$ and the time since this value was updated. An aircraft can always ask to be registered in or deleted from another aircraft's table in order to become neighbors or break the relationship. The second process is executed periodically with a frequency of $2$Hz. It calculates $^iu_r$ in (\ref{eq: urpapa}) from all the updated data not older than $2$ seconds, so we avoid situations like an aircraft that abandoned the formation but it continues registered in its neighbors table. Then the aircraft updates the radius to be tracked by the guidance vector field. Finally, the aircraft updates its own $\theta_i$ and transmit it to its neighbors if and only if the GPS is reliable, e.g., it has \emph{3D Fix}. Consequently, if the aircraft does not update its neighbors' tables, then it will not be taken into account in the formation after the timeout. This process is summarized in Algorithm \ref{alg: al}.

\begin{table}
	\centering
    \begin{tabular}{ | c | c | c |}
    \hline
	$\mathcal{N}_i$ & $\theta_{j\in\mathcal{N}_i}$ & Timeout(ms) \\ \hline
    2 & $\pi$ & 1042 \\ \hline
    3 & $0.4\pi$ & 426 \\ \hline
    \end{tabular}
	\caption{Example of table with neighbor's information.}
	\label{table: t}
\end{table}

\begin{algorithm}
 \KwResult{Rendezvous of aircraft $i$ with its neighbors.}
 \KwData{$\theta_i$ and Table \ref{table: t}.}
  \While{Formation Control == True}{
  $^iu_r = 0$\;
	\For{Visit all the rows of the table}{
  \If{Timeout is not reached}{
	  $^iu_r = {^iu_r} + (\theta_i - \theta_j)$ \;
   }
 }
	Set radius $r^2 + {^iu_r}$ in the guidance vector field\;
	\If{GPS is reliable}{
	Transmit $\theta_i$ to $\mathcal{N}_i$\;
	}
	}
 \caption{Algorithm executed at aircraft $i$ once the formation control is activated. Note that is not only distributed but it does not need all the information from all the neighbors to be executed.}
 \label{alg: al}
\end{algorithm}

\section{IMAV formation flight parade}
\label{sec: ff}
The three employed aircraft are custom 1.5m wingspan flying-wings equipped with the following electronics:
\begin{itemize}
	\item 2x servos SAVOX SH-0257MG
	\item ESC Flyduino KISS 18A v1.2
	\item Brushless motor T-Motor MT2208-18 1100KV
	\item Propeller 8x6
	\item 3S Battery pack built from Panasonic NCR18650B (up to 40min endurance, 2h+ with 2x3S)
	\item Apogee board autopilot\footnote{https://wiki.paparazziuav.org/wiki/Apogee/v1.00}
	\item Futaba receiver R6303SB
	\item Futaba transmitter FAAST
	\item 2x XBEE Pro S1 (on board and on ground)
	\item GPS M8
	\item Ground Control Station: Laptop running Paparazzi on Ubuntu 17.04
\end{itemize}

\begin{figure}
\centering
	\includegraphics[width=1\columnwidth]{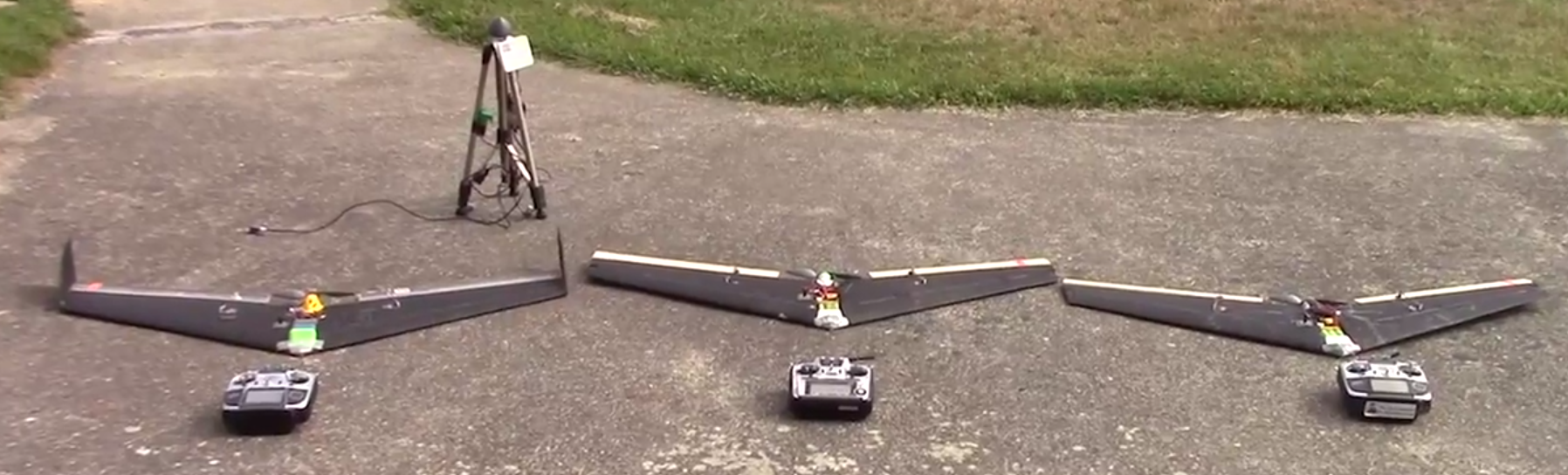}
	\includegraphics[width=1\columnwidth]{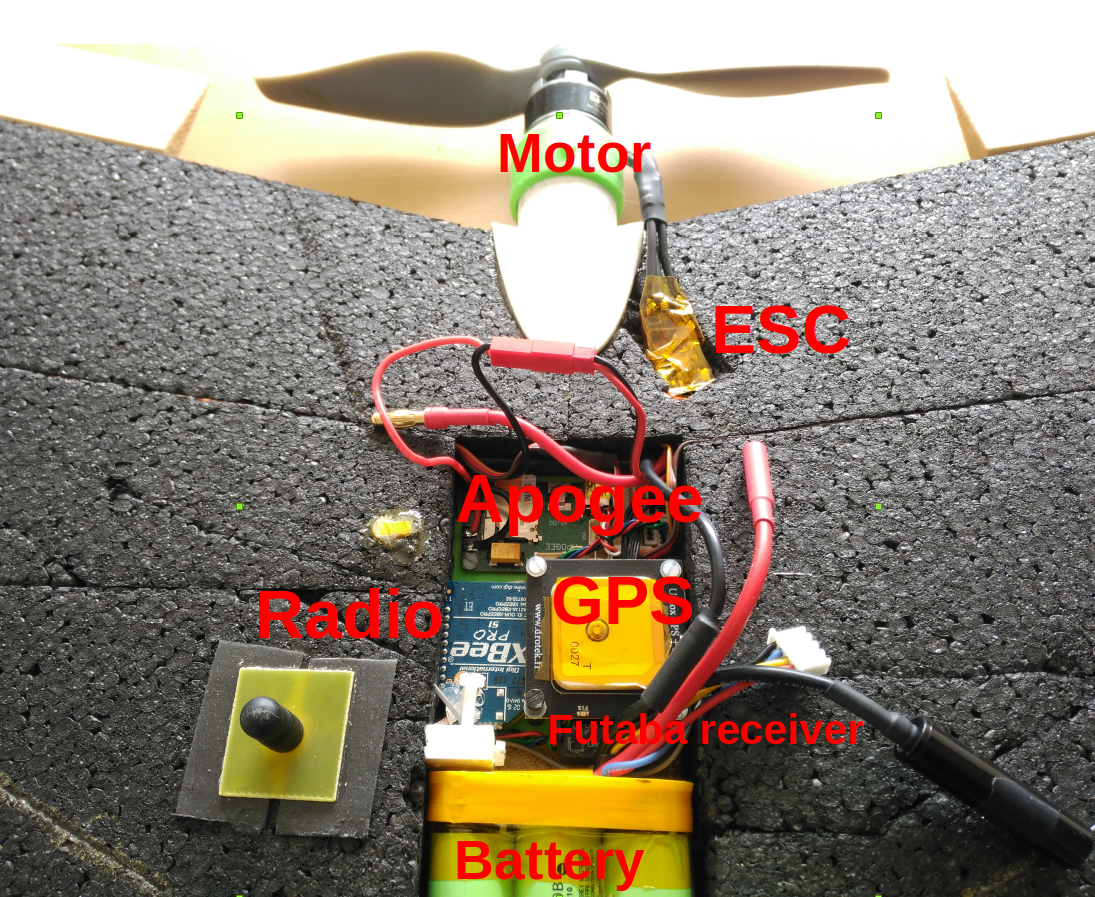}
	\caption{The three custom flying-wings employed for the formation flight parade equipped with Paparazzi autopilot.}
	\label{fig: wings}
\end{figure}

\begin{figure}[h]
\centering
	\includegraphics[width=1\columnwidth]{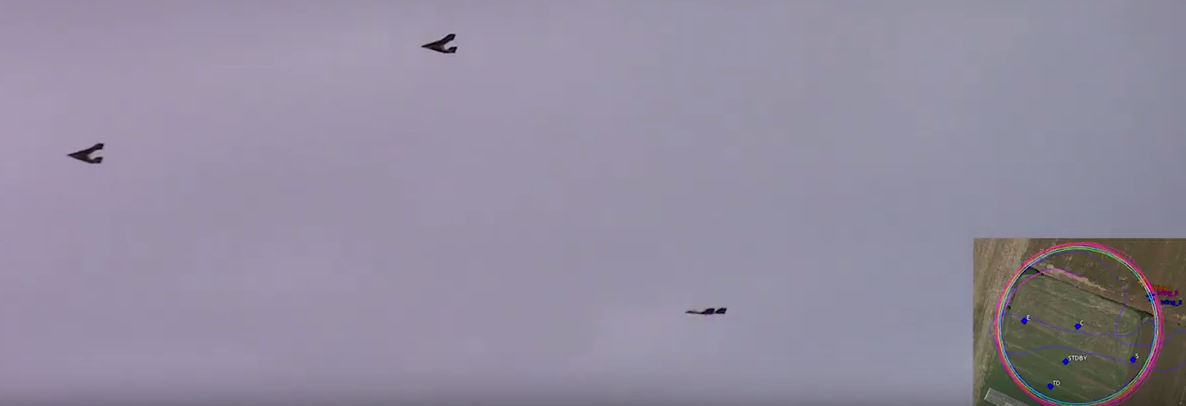}
	\caption{Picture taken from an actual formation flight executing the algorithm described in this paper.}
	\label{fig: ff}
\end{figure}

\begin{figure*}
\centering
\begin{subfigure}{.49\textwidth}
  \centering
  \includegraphics[width=\linewidth]{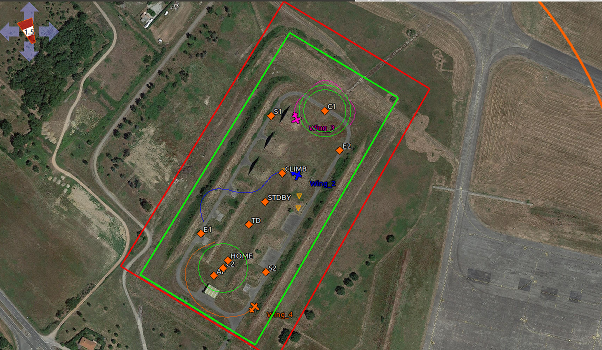}
  \caption{t = $51$secs}
\end{subfigure}
\begin{subfigure}{.49\textwidth}
  \centering
  \includegraphics[width=\linewidth]{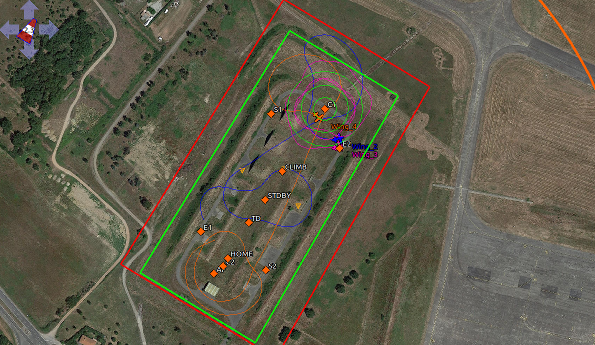}
   \caption{t = $67$secs}
\end{subfigure}
\begin{subfigure}{.49\textwidth}
  \centering
  \includegraphics[width=\linewidth]{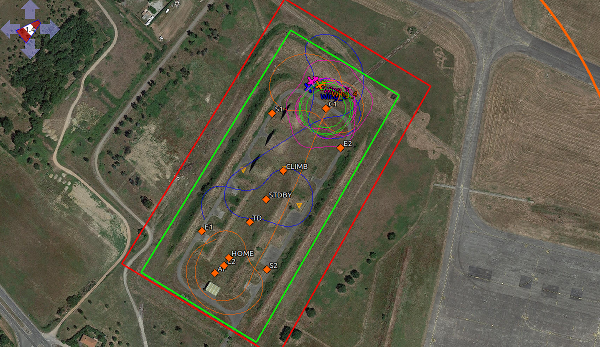}
   \caption{t = $78$secs}
\end{subfigure}
\begin{subfigure}{.49\textwidth}
  \centering
  \includegraphics[width=\linewidth]{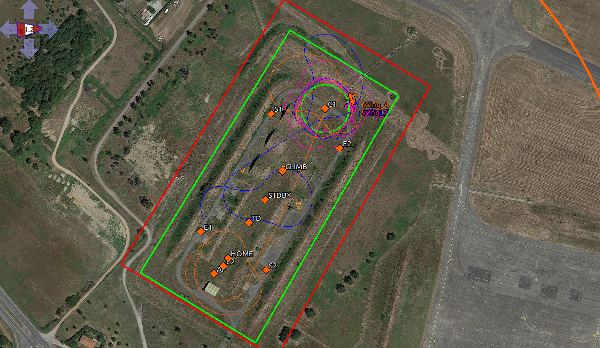}
  \caption{t = $95$secs} 
\end{subfigure}
\begin{subfigure}{.49\textwidth}
  \centering
  \includegraphics[width=\linewidth]{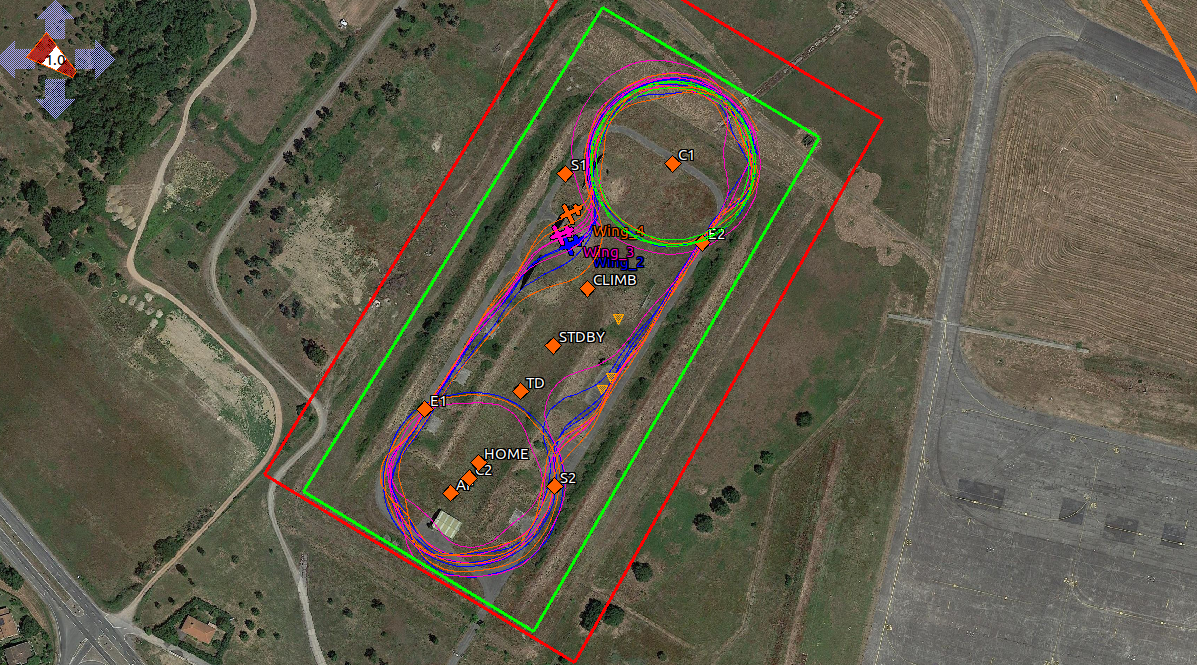}
	\caption{}
\end{subfigure}
\begin{subfigure}{.49\textwidth}
  \centering
  \includegraphics[width=\linewidth]{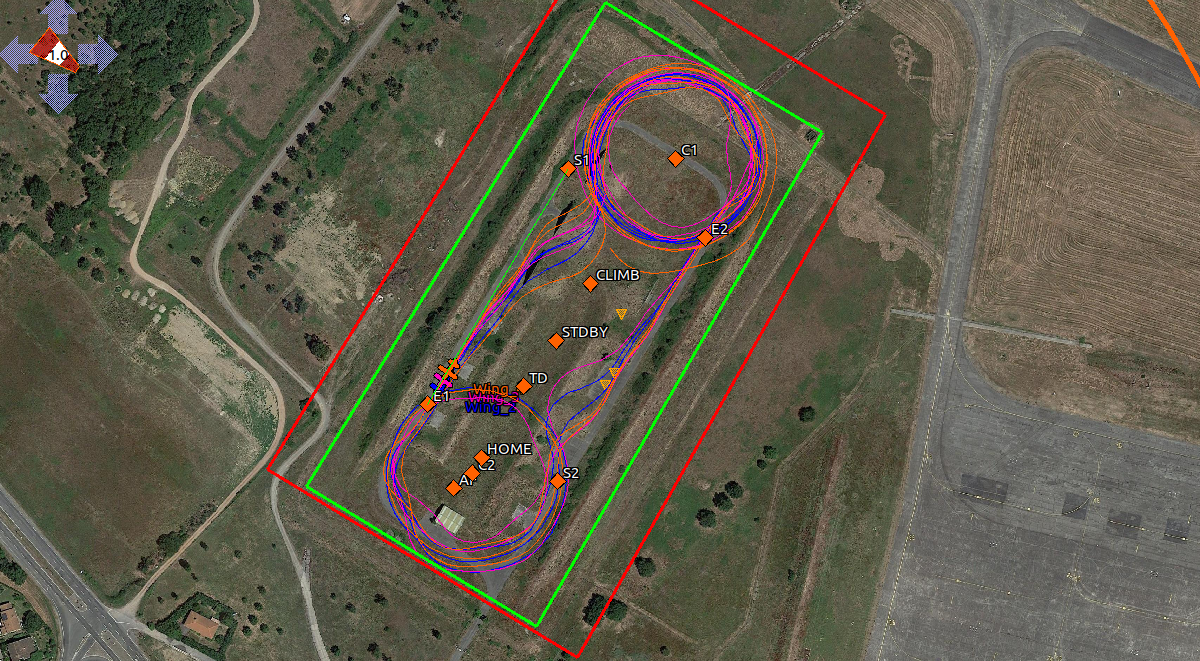}
   \caption{}
\end{subfigure}%
	\caption{Screenshots from the Paparazzi ground control station showing the evolution of the circular formation. The first screenshot starts with the aircraft at arbitrary positions within the allowed flying area for the IMAV outdoor competition. The convergence to a synchronized formation flight takes place in around thirty seconds without leaving the allowed flying area. The circular synchronization can be used for \emph{launching} the aircraft to travel together over the same segment.}
\label{fig: ppz}
\end{figure*}

We perform a formation flight simulation of three aircraft planned for the parade show in IMAV2017. For real experiment results like the one from the Figure \ref{fig: ff} we refer to \cite{iros2017}. Although the available space for maneuvering is tight, the simulation shows that it is possible to synchronize the aircraft flying at 11m/s (ground speed) in circumferences of radius 30 meters as it is shown in Figure \ref{fig: ppz}. 

\section{Conclusions}
\label{sec: con}
In this paper we have presented a fully distributed implementation of a flight formation controller for fixed-wing aircraft in the open-source autopilot Paparazzi. We manipulate guidance vector fields in order to control the inter-vehicle positions by changing the radius of the circumference to be eventually tracked. We finally show the effectiveness of the proposed strategies based on the rules for the IMAV2017 drone parade outdoor competition.


\bibliographystyle{unsrtnat}
\bibliography{imav_bibliography}

\appendix
\newcommand{\appsection}[1]{\let\oldthesection\thesection
  \renewcommand{\thesection}{Appendix \oldthesection:}
  \section{#1}\let\thesection\oldthesection}

\end{document}